\documentclass[journal,10pt]{IEEEtran}
\ifCLASSINFOpdf
\else
\fi
\usepackage{url}
\usepackage{framed,multirow}

\usepackage{amssymb}
\usepackage{latexsym}
\usepackage{bbm}
\usepackage{graphicx}
\usepackage{amsmath}
\usepackage{subfigure}
\usepackage[english]{babel}
\usepackage{booktabs}

\usepackage{url}
\usepackage{xcolor}

\def\eg{\emph{e.g. }}

\def\etal{\emph{et al. }}
\def\ie{\emph{i.e. }}

\begin{document}
\title{Learning to Exploit the Prior Network Knowledge \\ for Weakly-Supervised Semantic Segmentation}

\author{Carolina Redondo-Cabrera, Marcos Baptista-R\'ios and Roberto J. L\'opez-Sastre% <-this % stops a space
\thanks{Carolina Redondo-Cabrera, Marcos Baptista-R\'ios and Roberto J. L\'opez-Sastre are with the Research group GRAM of the Department of Signal Theory and Communications, 
University of Alcal\'a, Alcal\'a de Henares, 28805, Spain. e-mail: carolina.redondoc@edu.uah.es, marcos.baptista@uah.es, robertoj.lopez@uah.es}}


\markboth{IEEE Transactions on Image Processing,~Vol.~xx, No.~xx, xx~xx}%
{Redondo-Cabrera \etal }



\maketitle

\begin{abstract}
Training a Convolutional Neural Network (CNN) for semantic segmentation typically requires to collect a large amount of accurate pixel-level annotations, a hard and expensive task. In contrast, simple image tags are easier to gather. With this paper we introduce a novel weakly-supervised semantic segmentation model able to learn from image labels, and just image labels. Our model uses the prior knowledge of a network trained for image recognition, employing these image annotations as an attention mechanism to identify semantic regions in the images. We then present a methodology that builds accurate class-specific segmentation masks from these regions, where neither external objectness nor saliency algorithms are required. We describe how to incorporate this mask generation strategy into a fully end-to-end trainable process where the network jointly learns to classify and segment images. Our experiments on PASCAL VOC 2012 dataset show that exploiting these generated class-specific masks in conjunction with our novel end-to-end learning process outperforms several recent weakly-supervised semantic segmentation methods that use image tags only, and even some models that leverage additional supervision or training data.
\end{abstract}


% \fi
\IEEEpeerreviewmaketitle

\section{Introduction}
\label{intro}

Semantic image segmentation results of fundamental importance in a wide variety of computer vision tasks, such as scene understanding or image retrieval, but it comes at the cost of requiring pixel-wise labelling to generate training data. To collect a large amount of accurate pixel-level annotations is a hard and time consuming task. Therefore, this creation of training data has become one of the bottlenecks for the progress of semantic segmentation methods. Fortunately, a series of Convolution Neural Networks (CNNs)-based approaches for semantic segmentation, which rely on weaker forms of annotation, have emerged as a solution to this problem. They are known as weakly-supervised semantic segmentation solutions, \eg \cite{Bearman2016, Meng2017, Oh2017, Papandreou2015, Pathak2015, Pathak2014, Pinheiro2015, Qi2016, Saleh2016, Wei2017, Wei2017b, Wei2016,Zhang2014}.

A particularly appealing setting is defined by those models that are able to perform a semantic segmentation using only image-level labels to indicate the presence or absence of the classes of interest, which are rather inexpensive attributes to annotate and thus more common in practice (\eg Flickr \cite{Huiskes2008}). All these models share a common problem: how can we accurately assign image-level labels to corresponding pixels of training images such that the CNN-based models can learn to perform a semantic segmentation?

Some approaches (\eg \cite{Pathak2015, Pathak2014}) propose to use the available image labels as a constraint on the output segmentation mask. If an image tag is absent, no pixel in the image should take that label; if an image label is present, the segmentation must contain the label at least in one pixel. However, these weakly-supervised segmentation algorithms typically yield poor localization accuracy, because, for instance, the objects of interest are rarely single pixel. To overcome this weakness, some approaches have proposed to exploit an objectness prior (\eg \cite{Bearman2016, Pinheiro2015, Qi2016, Wei2016}), with the drawback that the final segmentation heavily depends on the success of this external objectness module, which, in practice, only produces a coarse heat map which does not accurately determine the location and shape of the objects. Others substitute the objectness module with a saliency step (\eg \cite{Oh2017, Wei2017, Wei2017b}), which focuses the attention of the segmentation on regions that stand out in the image, which is not the case for \emph{all} the foreground objects in a scene.

\begin{figure*}[t]
\centering
\includegraphics[width=\linewidth]{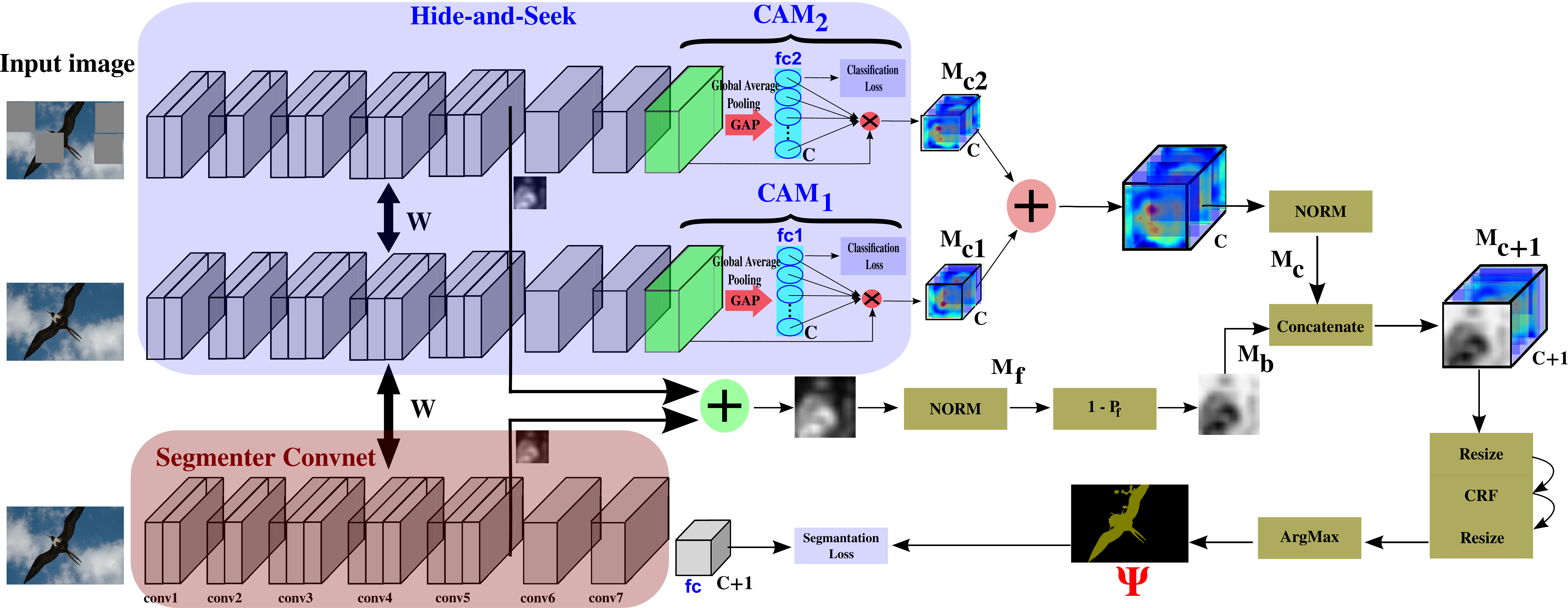}
\caption{Overview of the proposed weakly-supervised segmentation architecture. Our approach consists of two components: 1) the Hide-and-Seek module, whose aim consists in generating the class activation masks ($\Psi$); and 2) the Segmenter network, which learns to segment the images according to the activation maps.}
\label{fig:approach_global}
\end{figure*}

Motivated by all these shortcomings, we introduce a CNN-based architecture for weakly-supervised semantic segmentation, which can be trained fully end-to-end, hence not requiring any external tool for saliency or objectness prediction, and where the object position information is jointly extracted when the network is trained for image classification. It is well known that CNNs have remarkable object localization ability, despite being trained only on image level labels for classification and not for object detection, see \cite{Oquab2014, Zhou2016}, for instance. We leverage this capability to present our novel deep learning architecture for weakly-supervised semantic segmentation, shown in Figure \ref{fig:approach_global}, where the main contributions can be summarized as follows:

\begin{enumerate}
 \item We propose to integrate into the semantic segmentation architecture, \ie the \emph{segmenter} network in Figure\ref{fig:approach_global}, a novel procedure able to generate accurate class-specific activation masks to be used during learning. Technically, this procedure consists in a modification of the Global Average Pooling (GAP) layer combined with the Class Activation Mapping (CAM) technique \cite{Zhou2016}, which allows our network, in a single forward-pass, to both classify the image and localize class-specific segmentation masks.
 
 \item We introduce, in Section \ref{sec:am}, the Hide-and-Seek strategy, where we propose to combine two siamese CAM modules so as to recover activation masks covering the full object extents. As it is shown in Figure \ref{fig:approach_global}, our idea consists in randomly \emph{hiding} patches in a training image, forcing the second CAM network to \emph{seek} other relevant parts. This procedure with two siamese CAMs enforces the network to give attention not only to the most discriminative parts, but also to others which are less discriminative but fundamental to build better segmentation masks.
 
 \item To learn our whole model, we propose an end-to-end training process, which is based on a switching mechanism to alternate between two loss functions, allowing our architecture to jointly solve the segmentation and classification tasks. See Section \ref{sec:sl}.
 
 \item Section \ref{sec:test} shows how we propose to modify the standard inference process, implementing a filtering stage which leverages the knowledge of the Hide-and-Seek strategy to weight the importance of the class labels in the final segmentation. 
 
 \item Our solutions are learned from image labels, not requiring any external tool for saliency or objectness predictions. This allows us to offer a compact architecture, learned fully end-to-end, where the segmentation and classification tasks intertwine.
 
 \item In Section \ref{sec:exp} , we demonstrate the benefits of our approach on PASCAL VOC 2012 \cite{voc2012}, which is the most popular dataset for weakly-supervised semantic segmentation. Our experiments show that our model consistently outperforms several methods that use image tags only, and even some models that leverage additional supervision or training data.
\end{enumerate}

The remainder of this paper is organized as follows. Section \ref{sec:relatedwork} reviews the previous works found in the literature on the weakly-supervised semantic segmentation task. We formally define and discuss the individual components of our model in Section \ref{sec:app}. In Section \ref{sec:exp}, we provide further insights by discussing and evaluating the effect of each of our contributions separately in different experiments. We also compare the results achieved by our weakly-supervised semantic segmentation approach with the previous models and the state of the art. We conclude in Section \ref{sec:conclusion}.

\section{Related Work}
\label{sec:relatedwork}

The last years have seen a tremendous interest on weakly-supervised semantic segmentation, because this technique alleviates the painstaking process of manually generating pixel-level training annotations. To reduce the burden of this type of annotation, several strategies of weak supervision based on CNNs have been applied to the semantic segmentation task with great success. For instance, Papandreou \etal \cite{Papandreou2015} use annotated bounding boxes, and Lin \etal \cite{Lin2016} propose to employ scribbles.

Some recent works propose to train the solutions by just using image labels \cite{Bearman2016, Oh2017, Papandreou2015, Pathak2015, Pathak2014, Pinheiro2015,Qi2016, Saleh2016, Wei2017, Wei2017b, Wei2016, Wei2016b, Kolesnikov2016, Li2018, Wang2018, Huang2018, Wei2018, Kim2017}, which is probably the simplest form of supervision for training semantic segmentation models. These works can be classified in two groups: i) those that explore the use of external object saliency \cite{Oh2017, Wei2017, Wei2017b, Wang2018} or objectness priors \cite{Bearman2016, Pinheiro2015, Qi2016, Wei2016, Wei2016b} to build the rough segmentation masks to be used during learning; and ii) those that propose a closed solution, where the CNN segmentator is trained fully end-to-end, without any \emph{external} aid, \eg  \cite{Papandreou2015, Pathak2015, Pathak2014, Pinheiro2015, Saleh2016, Kolesnikov2016, Li2018, Huang2018, Wei2018, Kim2017}. Our model belongs to the second group.

Within this second group, the work of Pathak \etal \cite{Pathak2014} constitutes the first method to consider to fine-tune a pre-trained CNN, using only image-level tags, for a semantic segmentation task. Their approach relies on a Multiple Instance Learning (MIL) loss to account for image tags during training. While their model improves the segmentation accuracy of a naive baseline, the accuracy remains relatively low, due to the fact that no other priors than image tags are employed. Pinheiro \etal \cite{Pinheiro2015} also propose to use a MIL-based strategy to train models for semantic segmentation. They adopt an alternative training procedure, based on the Expectation-Maximization algorithm, to dynamically predict semantic foreground and background pixels. Their training procedure significantly increases the accuracy. Moreover, their work shows that it is possible to further improve this accuracy by introducing a stronger supervision, such as labeled bounding boxes for the objects. Importantly, however, their training procedure is dataset-dependent and somehow not trivial to be applied to a new dataset. Furthermore, their results remain inaccurate in terms of object localization/segmentation. In \cite{Pathak2015}, weakly-supervised semantic segmentation is formulated as a constrained optimization problem, and an additional prior modeling for the size of objects is introduced. This prior relies on thresholds determining the percentage of the image area that certain object categories can occupy, which again is dataset-dependent. Furthermore, as in \cite{Papandreou2015}, the resulting method does not exploit any information about the location of the objects, and thus yields poor localization accuracy. More recent works try to increase the quality of the object localization maps by: a) using dilated convolutions and varying their dilation rates \cite{Wei2018}; b) integrating a seeded region growing technique into the segmentation network \cite{Huang2018}; or c) proposing a novel framework able to provide a self-guidance mechanism on the attention map generation \cite{Li2018}.

Increasing this accuracy typically requires the application of saliency and/or objectness models. In particular, Bearman \etal \cite{Bearman2016} directly incorporate an objectness prior obtained by applying the algorithm in \cite{Alexe2012}. \cite{Pinheiro2015, Qi2016, Wei2016} employ object proposals obtained by the Multiscale Combinatorial Grouping (MCG) algorithm. Note that the MCG must be pre-trained with pixel-level annotations, therefore these three approaches cannot be considered in the \emph{fully} weakly-supervised segmenters group, they inherently are using a stronger supervision.

Oh \etal \cite{Oh2017} exploit the fact that CNN-based image classifiers are sensitive to discriminative areas of an image, and they propose to combine this knowledge with a saliency algorithm. They use the saliency model as an additional source of information and hereby exploit prior knowledge on the object extent and image statistics. In \cite{Wei2017}, an iterative adversarial erasing approach based on CNN image classifiers is presented. In order to expand and refine the object of interest regions, this approach drives the classification networks to sequentially discover new object localizations by erasing the current mined regions. Then, to further enhance the quality of the discovered regions, they use a saliency detection technology to produce the segmentation mask for the training images. Wei \etal \cite{Wei2017b} introduce a learning pipeline, in which an initial segmentation model is trained with simple images using just saliency maps for supervision.

In this paper we move towards a different direction, betting on solutions trained in a end-to-end fashion, \ie without a pipeline with external tools with complex learning procedures. We leverage the recent works \cite{Oquab2014,Zhou2016}, where it is described how, within its hidden layers, a CNN network trained for image classification only, have already learned to focus on generic foreground objects. In other words, a CNN network trained for object recognition is able to both classify the image and localize class-specific object regions in a single forward pass. 

Therefore, we present a weakly-supervised semantic segmentation solution, where our network simultaneously learns to generate a pool of multi-class attention masks using the mechanisms described in \cite{Oquab2014,Zhou2016}, and to perform the final segmentation according to these masks. All this, without any external aid apart from the image-level annotations.

The closest work to our approach is the one of Saleh \etal \cite{Saleh2016}. They also make use of the object localization information extracted directly from the classification network itself, as a cue to improve the semantic segmentation. However, there are important differences between their work and our solution. First, in \cite{Saleh2016} a mechanism to build a binary mask (foreground vs. background), using the network features, is described. We instead propose a network design which is able to generate class specific masks, using the activations from the features learned by a set of siamese networks. To the best of our knowledge, our approach is the first one to extract this information directly from the hidden layer activations of a classification network, and employ the resulting masks as localization cues for weakly-supervised semantic segmentation. Second, their learning process is completely different from ours. They incorporate the resulting foreground/background masks in their CNN network through a single semantic segmentation loss function based on \cite{Pathak2015}, following the optimization procedure Log-Sum-Log trick of \cite{Pinheiro2015}. We train our CNN architecture by jointly minimizing a weakly-supervised semantic segmentation loss and a couple of classification losses. Moreover, we use different semantic segmentation losses, and propose a switching loss learning procedure which reports the best results in our experiments.

Finally, we find the recent works \cite{Singh2017} and \cite{Zhang2018} on the different problem of weakly supervised object localization, that propose to use similar learning strategies to our Hide-and-Seek approach. In \cite{Singh2017}, a hide and seek module is also introduced to train the original CAM of \cite{Zhou2016} using random hidden image patches to increase the precision of the class activation maps. Instead, we use a Siamese deep network architecture that allows our approach to simultaneously learn the semantic segmentation module and the object localizers. To have a Siamese based model brings some important benefits: i) we can continuously refine the activation maps when the whole image is visible and when the patches are hidden; ii) the weight sharing mechanism behind a Siamese architecture allows to deploy an end-to-end optimization where all the tasks (object localization and semantic segmentation) are interconnected. Moreover, while in \cite{Singh2017} only changes to the input image are applied to train the original CAM \cite{Zhou2016}, we perform some modifications into the CAM architecture itself. In \cite{Zhou2016} and \cite{Singh2017} the global average pooling is done after the conv5 layer. We, instead, apply the global average pooling from the last convolutional layer, \ie conv7. This change is justified in the experiments, which reveal that it results beneficial for the semantic segmentation task. With respect to the other work, \ie \cite{Zhang2018}, we find the following significant differences. Zhang \etal \cite{Zhang2018} introduce the Adversarial Complementary Learning (ACL) for the problem of object localization. First, ACL technique needs to learn \emph{two different} classifiers (training the two adversarial branches), while in our Siamese architecture \emph{all the parameters are shared} by the two CAMs and the Segmenter network, an aspect that considerably reduces the number of parameters of the model. Second, in the ACL pipeline the adversarial classifier needs the input of the first classifier to erase the regions of the images. In contrast, our two CAM modules in the Hide-and-Seek block are in fact simultaneously learned. Overall, while \cite{Singh2017} and \cite{Zhang2018} have been designed to just solve the problem of object localization, our Siamese architecture allows us to adopt an end-to-end learning methodology to simultaneously solve the object localization and the semantic segmentation tasks.

\section{Weakly-Supervised Semantic Segmentation from Image-level Labels}
\label{sec:app}

In this section, we introduce our approach to weakly-supervised semantic segmentation. See Figure \ref{fig:approach_global} for the big picture of the proposed architecture. We present a closed model, without external aids or components, which is learned totally end-to-end. Our input consists of images with their associated labels, indicating the object categories they contain. These images are first used by our Hide-and-Seek module, in charge of learning to produce rough segmentation masks. We then propose a weakly-supervised learning algorithm that
leverages these masks, understanding them as the attention mechanism for our segmenter network, which should control the semantic segmentation precision.

Section \ref{sec:am} details how our approach extracts masks directly from a CNN network trained for object class recognition. Then, in Section \ref{sec:sl}, we present our architecture for semantic segmentation and we describe a new learning process to train our full model. The key idea of our learning process consists in that during training the proposed model switches between two loss functions, depending on how noisy the segmentation masks provided by the Hide-and-Seek module are.

\subsection{Hide-and-Seek: Finding Good Segmentation Masks via Class Activation Maps}
\label{sec:am}

There has been a recent burst of techniques for localizing objects from a CNN based classifier, \eg \cite{Oquab2015, Simonyan2013, Springenberg2014, Zhang2016, Zhou2016}. A strategy for image-level supervised localization based on CNNs is to produce activation maps (score maps) for each specific object category, and select or extract some representative activation values. Some approaches rely on image gradients from trained classifiers \cite{Simonyan2013, Springenberg2014, Zhang2016}, while others (\eg \cite{Oquab2015, Zhou2016}) propose to apply the Global Average Pooling (GAP) strategy of \cite{Lin2013} on the last convolution layer. The generated representations are then used as inputs to a fully-connected layer for predicting image class labels. Ultimately, a CNN based image classifier can therefore be also thought of as an object localizer: the classification brings the location of objects for free.

Ideally, one should simply incorporate this capability to generate pixel level information from just image level tags, in order to train semantic segmentation solutions. But, unfortunately, a direct application of these learned masks does not produce an acceptable accuracy of the final segmenter. First, because the masks are too rough. And second, because all these techniques do not necessarily capture the whole object but just some parts that are considered as discriminative for the convolutional layers. 

To solve these weaknesses, we propose to train a siamese CNN architecture by: a) jointly minimizing both a segmentation and a classification loss; and b) following our Hide-and-Seek methodology.

Technically, we propose, as it is depicted in Figure \ref{fig:approach_global}, a siamese architecture, which consists of three CNNs sharing their weights. Two belong to the Hide-and-Seek module, and one is for the Segmenter. Concretely, for these networks, we follow the VGG-16 design \cite{Simonyan2014}, pre-trained on the ILSVRC 2012 dataset \cite{Deng2009}. 

We now focus on the two networks of the Hide-and-Seek step. The classification loss we choose is the squared label prediction loss, as suggested by \cite{Wei2016}. This way, we can generate a sort of \emph{heat map} for each image-level label using the Classification Activation Maps (CAM) strategy \cite{Zhou2016}. Let $f_k(x,y)$ represent the activation of unit $k$ at spatial location $(x,y)$ in the last convolutional layer. Then, for unit $k$, the result of performing global average pooling, $F^k$, is $\sum_{x,y} f_k(x,y)$. Thus, the predicted object score for class $c$, $s_{o_c}$, which is the input to our squared label prediction loss, can be written as follows,

\begin{equation}
\label{score}
 s_{o_c} = \sum_{k}\omega_{k}^{c}F^k = \sum_{x,y} \sum_{k}\omega_{k}^{c} f_k(x,y), 
\end{equation}
where $\omega_k^c$ is the weight corresponding to class $c$ for unit $k$. Essentially, $\omega_k^c$ indicates the importance of the unit $k$ for the object category $c$, with $c = \{1, \dots, C\}$. $C$ indicates the number of object categories in the training set. Note that, as it is suggested in \cite{Zhou2016}, we ignore the bias term.

To generate a class activation map for class $c$, $M_c$, one can use these weights in a linear combination of the activations of the units in the last convolution layer, following
\begin{equation}
\label{map}
 M_c(x,y)=\sum_k \omega_k^c f_k(x,y).
\end{equation}

Thus, $s_{o_c} = \sum_{x,y} M_c (x, y)$, where $M_c$ directly indicates the importance of the activation at position $(x,y)$ leading to the classification of an image to category $c$.

These activation maps suffer from two main drawbacks, like we just said. First, they only roughly match the shape of the object, yielding an inaccurate localization of the object's boundary. Second, they only focus on object parts which have been considered as discriminative during the classification optimization. Certainly, they do not tend to cover the whole object and thus complicate the semantic segmentation task.

To overcome these limitations, we propose to use two siamese CAM modules, being this the main idea of our Hide-and-Seek module. As it is shown in Figure \ref{fig:approach_global}, we propose to train a second CAM network but with a different input. Technically, we let this extra network to learn the CAMs using input images where we have \emph{randomly} hidden some regions. This procedure forces our architecture to seek other relevant parts of the objects, incorporating them to the CAMs. Ultimately, our strategy works as a mechanism to draw the attention of the networks to more and more parts of the objects, building CAMs that are actually able to localize a larger extent of the objects.

We first divide the input images using a grid of non-overlapping regions. These regions are randomly \emph{hidden} in each iteration. Therefore, it might happen that an object of interest is completely hidden, or that the object is split (\ie partially hidden). Neither of these situations is problematic. The former forces our architecture to focus on the context of the object that might be relevant for the segmentation task, although our background mask generator and the CRF refinement step will be in charge of determining the final contribution (see Figure \ref{fig:approach_global}). The latter allows our network to focus on a partial view of the object, hence focusing the attention of the network on the visible parts, which might not have been attended by the standard CAM.

In more detail, our Hide-and-Seek strategy is performed by dividing each training image into a grid of $P \times P$ non-overlapping patches. With probability $P_h$, each patch is then randomly \emph{hidden}, \ie its internal pixels are replaced by the mean pixel values of all the training images. This novel \emph{hidden} image is then given as input to the second CAM network to learn the corresponding image classifier. Note that the hidden patches change randomly across the different training epochs, which ideally forces the network to focus on different object parts during learning. Once the two input images are passed through our siamese CAM architecture, the activation maps ($M_{c1}$ and $M_{c2}$) are generated. Then, they are merged using an addition operation. Finally, the class activation maps are normalized between 0 and 1 obtaining the final $M_c$ class activation maps.

To be able to train the segmenter network we need the activation maps of each object category, but we also require an activation map for the \textit{background} class. To generate this background annotation is going to be fundamental for the accuracy of the final semantic segmentation. In practice, we use the CAMs learned by our two siamese CNNs. As our networks have learned to focus on the objects themselves, they should produce high activations values on the objects and their parts, and low activation responses for background-like regions, such as sky and roads. One can therefore obtain a \textit{foreground} map by computing the mean over the feature responses from each filter in the fifth convolutional layer (conv5) for each spatial location. As it is shown in Figure \ref{fig:approach_global}, since the input image is the same for the Segmenter CNN and the first CAM, we simply fuse the activation maps of the Segmenter convnet and the Hide-and-Seek second CAM ($CAM_2$) to obtain our \textit{foreground} mask ($M_f$). The fusion operation is performed by a simple element wise summation, followed by a normalization layer. So, the resulting activation \textit{foreground} map, $M_f$, can be thought of as a pixel wise foreground probability. Finally, if we denote $P_f(x, y)$ as the foreground probability at spatial location $(x, y)$ in $M_f$, one can obtain the \textit{background} activation map at spatial grid $(x, y)$ as $M_b(x, y)= 1 - P_f(x, y)$. With the \textit{background} activation map already computed, we concatenate it with the rest of class activation maps obtaining the ultimate set of activation maps ($M_{c+1}$), with $c+1$ layers. 

Once the core of the Hide-and-Seek process has been introduced, it is important to emphasize that our solution is trained fully end-to-end. The activation maps are jointly learned with the semantic segmentation network, this fact prevents us from having to rely on any external aid \cite{Oh2017, Wei2017, Wei2017b}. Moreover, we require neither pixel-level annotations \cite{Pinheiro2015, Qi2016, Wei2016}, nor object bounding boxes to train our model \cite{Papandreou2015, Pinheiro2015}. Image labels and just image labels.

As a final refinement stage of the activation maps, we make use of a fully connected Conditional Random Field (CRF) with higher-order terms in order to smooth out the segmentation maps computed by our Hide-and-Seek siamese CNNs architecture. At this point, we integrate into our system the fully connected CRF model of \cite{Krahenbuhl2011}. Let $z = \{z_i\}_{i = 1}^{W\cdot H}$ be the set of random variables, where $z_i$ encodes the pixel label, and $W$ and $H$ represent the width and the height, respectively, of the resulting activation map, after resizing it to the original image size. 

The model learns the joint distribution over all pixels with the following energy function:
\begin{equation}
 E(\textbf{z}) = \sum\limits_i \theta_i(z_i) + \sum\limits_{ij}\theta_{ij}(z_i, z_j),
\end{equation}
where $\textbf{z}$ encodes the label assignments for pixels. We use as unary potential, $\theta_i(z_i) = - \log P(z_i)$, where $P(z_i)$ is the label assignment probability at pixel $i$ as computed by our CNN architecture.

The pairwise potential $\theta_{ij}$ encodes the compatibility of a joint label assignment for two pixels. Following \cite{Krahenbuhl2011}, we define this pairwise term as a contrast-sensitive Potts model using two Gaussian kernels encoding color similarity and spatial smoothness. In particular, we use the following expression:
\begin{eqnarray}
\nonumber \theta_{ij} =   \mu(z_i,z_j) \Bigg [ \omega_1 \exp \left(- \frac{||p_i - p_j||^2}{2\sigma_{\alpha}^2} - \frac{||I_i - I_j||^2}{2\sigma_{\beta}^2}\right) \\
 +  \omega_2 \exp \left(- \frac{||p_i - p_j||^2}{2\sigma_{\gamma}^2} \right) \Bigg ], 
\end{eqnarray}
where $\mu(z_i, z_j ) = 1$ if $z_i \neq z_j$ , and zero otherwise, which means that only nodes with distinct labels are penalized. The remaining expression corresponds to two Gaussian kernels. The first one is a bilateral kernel that depends on both pixel positions (denoted as $p$) and RGB color (denoted as $I$). The second kernel only depends on pixel positions. The first kernel is an appearance kernel, inspired by the observation that nearby pixels with similar color are likely to be in the same class. The second kernel removes small isolated regions enforcing smoothness. Hyper-parameters $\sigma_{\alpha}$, $\sigma_{\beta}$ and $\sigma_{\gamma}$ control the scale of the Gaussian kernels. Parameters $\omega_1$ and $\omega_2$ determine the trade-off between the appearance and smoothness kernels. 

Finally, in order to attain our estimated segmentation mask ($\Psi$), which is used as supervision for the semantic segmentation model, we first resize the smoothed activation map back to the dimensions $w \times h$ of the feature or score map provided by the segmenter network. If we denote the predicted probability for the class $c$ of pixel $p_i$ as $p_i^c$, and the category set as $\mathcal{O}=\{0, 1, \ldots, C\}$, where $0$ indicates the background class, we can obtain the estimated label $l_{i}$ of each pixel $p_i$ in our segmentation mask ($\Psi$) as
\begin{equation}
 l_{i} = \arg\max\limits_{c \in \mathcal{O}} p_{i}^c.
\end{equation}

\subsection{Weakly-Supervised Learning by Switching Loss Functions}
\label{sec:sl}

%Description of the idea
We here introduce our weakly-supervised learning model. When the class label is available for every pixel during training, or, as in our case, when these labels are provided by the Hide-and-Seek module, the segmenter network can be trained by optimizing the sum of per-pixel cross-entropy terms. 

However, we cannot follow this simple learning pipeline. Since our CAM networks are learned simultaneously with the rest of our architecture, \ie we do not use any pretrained solution, the attention masks provided by them can be quite noisy, specially during the first iterations. Therefore, we propose to combine in the optimization a standard weakly-supervised loss for semantic segmentation, and a switching process for those iterations where the segmentation masks provided by our siamese CAM networks are not accurate.

We now first describe a loss for weakly-supervised learning of segmentation models, based on image labels only. Then, we show how to combine this loss with the traditional per pixel softmax cross-entropy loss along with the labels provided by our class activation maps.

Intuitively, given an image with different class labels, one would like to encourage the segmenter to assign image pixels to the observed classes in the image, while penalizing an assignment to unobserved classes. Formally, let $\mathcal{I}$ be the set of pixels in the feature map provided by the segmenter network. Let $L \subseteq \{ 0, 1, \ldots, C \}$ be the set of classes (including class 0, background) present in the image, and $\hat{L} \subseteq \{ 0, 1, \ldots, C \}$ be the set of classes not present in the image. Furthermore, let $s_{ic}$ be the segmenter score for pixel $i$ and class $c$. Then, the softmax probability of class $c$ at pixel position $i$ can be defined as follows, $S_{ic}= \exp(s_{ic})/\sum_{k=0}^{C}\exp(s_{ik})$. 

In this case, knowing only the set of classes present and not present in the image, one can train the model with the following cross-entropy loss:
\begin{equation}
\label{weak}
 \mathcal{L}_{weak} (S, L, \hat{L}) = - \frac{1}{|L|}\sum\limits_{c \in L} \log(S_{t_c}) - \frac{1}{|\hat{L}|} \sum\limits_{c \in \hat{L}} \log(1 - S_{t_c}),
\end{equation}
where $S_{t_c}$ represents a candidate score for each class present or not present in the image. The first part of Equation \ref{weak} is used in \cite{Pathak2014}. It encourages each present class to have a high probability on at least one pixel in the image. The second part has been introduced by \cite{Pathak2015}, corresponding to the fact that no pixels should have high probability for classes that are not present in the image.

Now that this weakly-supervised loss has been introduced, we detail how to combine it with the per pixel softmax cross-entropy loss, which is going to be used in conjunction with the pixel labels of the learned activation maps.

Given the estimated activation masks in $\Psi$, obtained by our siamese CAM networks, we have a mechanism to indicate whether pixel $i$ belongs to class $l_i$. Thus, the cross-entropy loss function proposed to train our segmentation network can be defined as
\begin{equation}
\label{smx}
 \mathcal{L}_{smx} (S, \Psi) = - \frac{1}{N}\sum\limits_{i \in \mathcal{I}}\log S_{il_i},
\end{equation}
where $N$ is the total number of pixels in the feature map.

Our learning process naturally optimizes the loss in Equation \ref{smx}. However, to learn only through this loss may not be the best choice. Like we just have explained, the segmentation masks provided by the Hide-and-Seek component can be imprecise in some iterations. Therefore, we propose to switch to the loss in Equation \ref{weak}, when we detect a noisy segmentation mask. In particular, we propose to apply a max pooling aggregation to Equation \ref{weak}. In our case, we take $S_{t_c}$ of Equation \ref{weak} as $S_{t_c} = \max\limits_{i \in \mathcal{I}} S_{ic}$. This encourages the model to increase the score of the pixel which is considered as the most important pixel for the image-level classification task. Note that we have chosen to apply this max pooling model, instead of the Log-Sum-Log trick \cite{Pinheiro2015, Saleh2016}. This decision has multiple benefits. First of all, the segmentation performance increases in all our experiments. Secondly, because specially in noisy iterations it is interesting to follow this pooling, which actually focuses the network learning only on one pixel, the one selected by the max operation, and not on multiple pixels with noisy label assignments.

\begin{figure*}[t]
\centering
\includegraphics[width=0.9\linewidth]{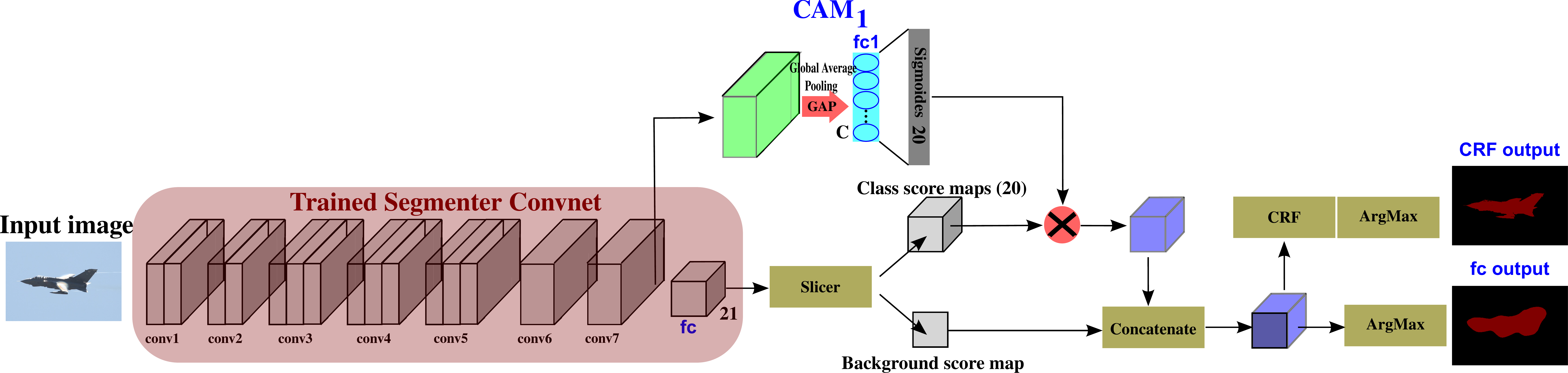}
\caption{Inference stage. We propose a filtering approach, which leverages the class activation knowledge of our Hide-and-Seek network to refine the final segmentation. Technically, we incorporate to the pipeline a classification branch which recovers the importance of each class in the test image. These scores per category are used to weight the output segmentation masks produced by the segmenter module.}
\label{fig:test_filter}
\end{figure*}

Technically, we have designed two mechanisms for the loss switching. The first one is based on a \textit{fixed early switching} model. In order to obtain appropriate activation maps with the Hide-and-Seek component, during the first  $n_i$ iterations, the segmenter is trained only using $\mathcal{L}_{weak}$, and the rest of iterations is trained by optimizing Equation \ref{smx}. 

The second scenario is based on an \textit{adaptive switching} mode throughout the learning process. Once the initial $n_i$ iterations have passed, the segmenter is optimized by $\mathcal{L}_{weak}$ only in noisy segmentation masks iterations. We consider a segmentation mask $\Psi$ as noisy if the number of pixels belonging to a different class from the background class is lower or equal than a certain threshold $\tau$. We consider this criterion appropriate, because we have detected that the Hide-and-Seek module, when it does not yield a correct result, is because it tends to classify all pixels in the background class. Note that this situation is particularly aggravated by the CRF model. These are precisely the masks that we want to avoid during the learning process, so that the segmentation network is not contaminated. Formally, we can write our adaptive switching loss function as

\begin{equation}
 \mathcal{L}_{adapt} = 
\begin{cases}
    \mathcal{L}_{weak},& \text{if } \sum\limits_{i \in \Psi}\mathbbm{1}[l_i \neq 0] \leq \tau\\
    \mathcal{L}_{smx},              & \text{otherwise},
\end{cases}
\end{equation}
being $\mathbbm{1}[l_i \neq 0]$ equal to $1$ if and only if the pixel label $l_i \in \Psi$ does not belong to class $0$. In practice, we set $\tau$ to $0$. We show the impact of both strategies in Section \ref{sec:switch_res}.

\subsection{Inference stage}
\label{sec:test}

Once the architecture has been learned, we are ready to enter into the inference mode. The segmenter network can be directly used to produce the semantic segmentation for test images. However, we propose to introduce a filtering step to further improve the performance of the segmentations. This filter is able to leverage the class activation maps knowledge, learned by the Hide-and-Seek module, to refine the final segmentation. 

As it is shown in Figure \ref{fig:test_filter}, we attach to the segmenter module a sort of classification branch, which recovers the importance of each class in the test image. Technically, we take the output of the last convolutional layer. This is passed trough the learned GAP and fully connected layer of the first CAM of the Hide-and-Seek module. This way the class activation maps for the test image can be recovered. We then use a sigmoid per class to obtain a score per category, which are used to weight the segmentations per class. This weighting mechanism regulates the importance of the different classes in the final segmentation, leveraging the knowledge of the Hide-and-Seek module. Therefore, those classes with low confidences are removed for obtaining the final segmentation mask. Our experiments show the benefits of this filtering process, which otherwise could not have been implemented without a learning mechanism as described in this article.

\section{Experiments}
\label{sec:exp}

In Sections \ref{sec:details} and \ref{sec:contributions}, we provide details of our implementation and an extensive evaluation of the contributions of our work, respectively. Finally, Section \ref{sec:state-art} compares our weakly-supervised semantic segmentation approach with the previous models trained in an end-to-end framework from image-level labels. We also compare our approach with other methods that rely on additional supervision, providing this way a detailed analysis with respect to the state of the art.

\subsection{Experimental Setup}
\label{sec:details}

\subsubsection{Implementation Details}

We adopt the architecture of DeepLab-CRF-LargeFOV \cite{Liangchieh2015} model for both the Hide-and-Seek and segmenter modules. We choose a VGG-16 \cite{Simonyan2014} based architecture, whose parameters are initialized learning the 1000-class classification task on the ILSVRC 2012 dataset \cite{Deng2009}. The last convolutional layer is initialized with zero-mean Gaussian noise, with a standard deviation of 0.1. We use a mini-batch of 1 image, and the parameters of the network are learned using stochastic gradient descent, with a learning rate of $10^{-4}$ for the first 40k iterations, and of $10^{-5}$ for the next 40k iterations. We use a momentum of 0.9 and fix the weight decay to 0.0005. We employ the original DeepLab code \cite{Liangchieh2015}, which is implemented based on the publicly available Caffe framework \cite{Jia2014}. Our codes can be downloaded from: \url{https://github.com/gramuah/weakly-supervised-segmentation}.

To train our model, input images are randomly cropped to patches with size $321 \times 321$. We directly feed the segmenter module with these patches, obtaining as output a feature map of width and height equal to $41 \times 41$. The Hide-and-Seek module needs these patches to be resized to $224 \times 224$. For the second CAM network, note that we first divide these patches into a grid with 16 regions of size $56 \times 56$. Each region is then hidden with a probability $P_h = 0.5$. We then take the new images with hidden regions, and we use them to feed this second CAM module. With this size of input, the resulting activation map ($M_c$) has a spatial resolution of $14 \times 14$. For the noise smoothing process, we set the CRF parameters following the setup described in \cite{Liangchieh2015}: $\omega_1 = 5$, $\omega_2 = 3$, $\sigma_{\alpha} = 50$, $\sigma_{\beta} = 10$ and $\sigma_{\gamma} = 3$. 

\subsubsection{Dataset and Evaluation Metrics}

In our experiments, we use the PASCAL VOC 2012 \cite{voc2012} dataset, which serves as a benchmark for most of the weakly-supervised semantic segmentation published papers \cite{Oh2017, Papandreou2015, Pathak2015, Pathak2014, Pinheiro2015, Saleh2016, Wei2017}. This dataset contains 20 object categories and one background category and 10582 training images (the original VOC 2012 training set and the additional data annotated by \cite{Hariharan2011}), 1449 validation images and 1456 test images. In our experiments, only image-level labels are used for training, and these image tags are obtained from the pixel-level annotation by simply listing the classes observed in each image. The performance is evaluated in terms of the Pixel-wise Classification Accuracy (PCA) of the predicted segmentation. We also report the mean Intersection over Union (mIoU) between the ground-truth and estimated segmentation masks.

\subsection{Ablation Study}
\label{sec:contributions}

We analyze here the main improvements associated to each of the proposed contributions. We start discussing in Section \ref{sec:gaps} about the different global average pooling (GAP) architectures. Section \ref{sec:num_cams} shows the influence in the performance of the Hide-and-Seek strategy. Then, in Section \ref{sec:crf_influence} we show the benefits of introducing a CRF based strategy during the learning process. Section \ref{sec:knowledge} demonstrates how the precision increases due to the filtering process introduced for the inference stage. Finally, we discuss about our switching loss process (Section \ref{sec:switch_res}) and the impact of training the model in an end-to-end fashion (Section \ref{sec:impact_endtoend}).

\subsubsection{GAP Architecture Analysis}
\label{sec:gaps}

Table \ref{table:result_voc_res} compares different GAP architectures. The first row corresponds to the original architecture described in \cite{Zhou2016}, where, for the VGG-16 network, the layers after convolution 5 (conv5) are removed and the GAP strategy is applied from this layer. The second row shows the results achieved by our proposal, which implements the global average pooling from the last convolution, \ie convolution 7 (conv7). For both architectures, we show the accuracy achieved by the model before and after the CRF utilized for post-processing during inference. Interestingly, when using a GAP strategy from conv7 instead of conv5, we see a significant improvement for all metrics. In more detail, in terms of mIoU, we gain $7.76\%$ and $8.85\%$ before and after CRF, respectively.

\begin{table}[t]
\caption{Comparing GAP architectures. Accuracy comparison on PASCAL VOC 2012 validation set.}
\label{table:result_voc_res}
\begin{center}
\scalebox{0.7}{
\begin{tabular}{l|cc|cc}
\toprule
\multicolumn{1}{c|}{Architectures} & \multicolumn{2}{c|}{\textbf{before CRF}} & \multicolumn{2}{c}{\textbf{after CRF}}\\
\multicolumn{1}{c|}{} & PCA ($\%$)  & mIoU ($\%$) & PCA ($\%$)  & mIoU ($\%$)\\
\midrule
GAP from conv5 & 70.92 & 26.27 & 73.60 & 27.65\\
%\midrule
GAP from conv7 & \textbf{72.92} & \textbf{34.03} & \textbf{75.98} & \textbf{36.50}\\
\bottomrule
\end{tabular}
}
\end{center}
\end{table}

\subsubsection{Hide-and-Seek module}
\label{sec:num_cams}

How important is the Hide-and-Seek module? In Table \ref{table:result_cam} we report the results achieved by our model using different number of siamese CAM networks. Note that when only one CAM network is used, no Hide-and-Seek strategy is actually implemented. It is when we incorporate two or more CAM networks, when we use the Hide-and-Seek process to focus the attention of these novel CAMs to images that contain randomly hidden patches. Again, we show the model performance before and after the CRF post-processing. As it can be seen, in terms of mIoU, the use of a pair of CAM networks improves the performance of an architecture with just one CAM and no Hide-and-Seek idea. The best result is obtained by introducing two siamese CAMs to the model.

\begin{table}[t]
\caption{Quantitative results with and without the Hide-and-Seek module. Accuracy comparison on PASCAL VOC 2012 validation set.}
\label{table:result_cam}
\begin{center}
\scalebox{0.7}{
\begin{tabular}{l|cc|cc}
\toprule
\multicolumn{1}{c|}{With or Without} & \multicolumn{2}{c|}{\textbf{before CRF}} & \multicolumn{2}{c}{\textbf{after CRF}}\\
\multicolumn{1}{c|}{Hide-and-Seek} & PCA ($\%$) & mIoU ($\%$) & PCA ($\%$) & mIoU ($\%$)\\
\midrule
Without (1 CAM) & 69.80 & 33.04 & 72.47 & 35.47\\
%\midrule
With (2 CAMs) & \textbf{72.92} & \textbf{34.03} & \textbf{75.98} & \textbf{36.50}\\
%\midrule
With (3 CAMs) & 72.03 & 33.79 & 75.76 & 36.04\\
\bottomrule
\end{tabular}
}
\end{center}
\end{table}

Demonstrated the benefits of the Hide-and-Seek mechanism, we explore now the influence of its parameters. Table \ref{table:cam_ana} summarizes the accuracy obtained by our model when the second CAM uses different number of hidden random patches per image. For this experiment, we use a fixed number of hidden random patches (rows: 1, 2, 3), and a random number which changes across different epochs (row 4). The best performance is obtained by the random strategy, which we believe forces the network to better learn the different parts of the objects, and therefore to obtain better class activation maps.

\begin{table}[t]
\caption{Hide-and-seek CAM analysis: Number of extracted hidden random patches. Accuracy comparison on PASCAL VOC 2012 validation set.}
\label{table:cam_ana}
\begin{center}
\scalebox{0.7}{
\begin{tabular}{l|cc|cc}
\toprule
\multicolumn{1}{c|}{Number of} & \multicolumn{2}{c|}{\textbf{before CRF}} & \multicolumn{2}{c}{\textbf{after CRF}}\\
\multicolumn{1}{c|}{random patches} & PCA ($\%$) & mIoU ($\%$) & PCA ($\%$) & mIoU ($\%$)\\
\midrule
4 & 72.64 & 33.13 & 75.63 & 35.80\\
%\midrule
5 & 73.03 & 33.34 & 76.09 & 36.18\\
%\midrule
6 & 72.59 & 33.18 & 75.60 & 35.90\\
%\midrule
random & \textbf{72.92} & \textbf{34.03} & \textbf{75.98} & \textbf{36.50}\\
\bottomrule
\end{tabular}
}
\end{center}
\end{table}

\subsubsection{Smoothing the Activation Maps}
\label{sec:crf_influence}

\begin{figure}[t]
\centering
\includegraphics[width=0.8\linewidth]{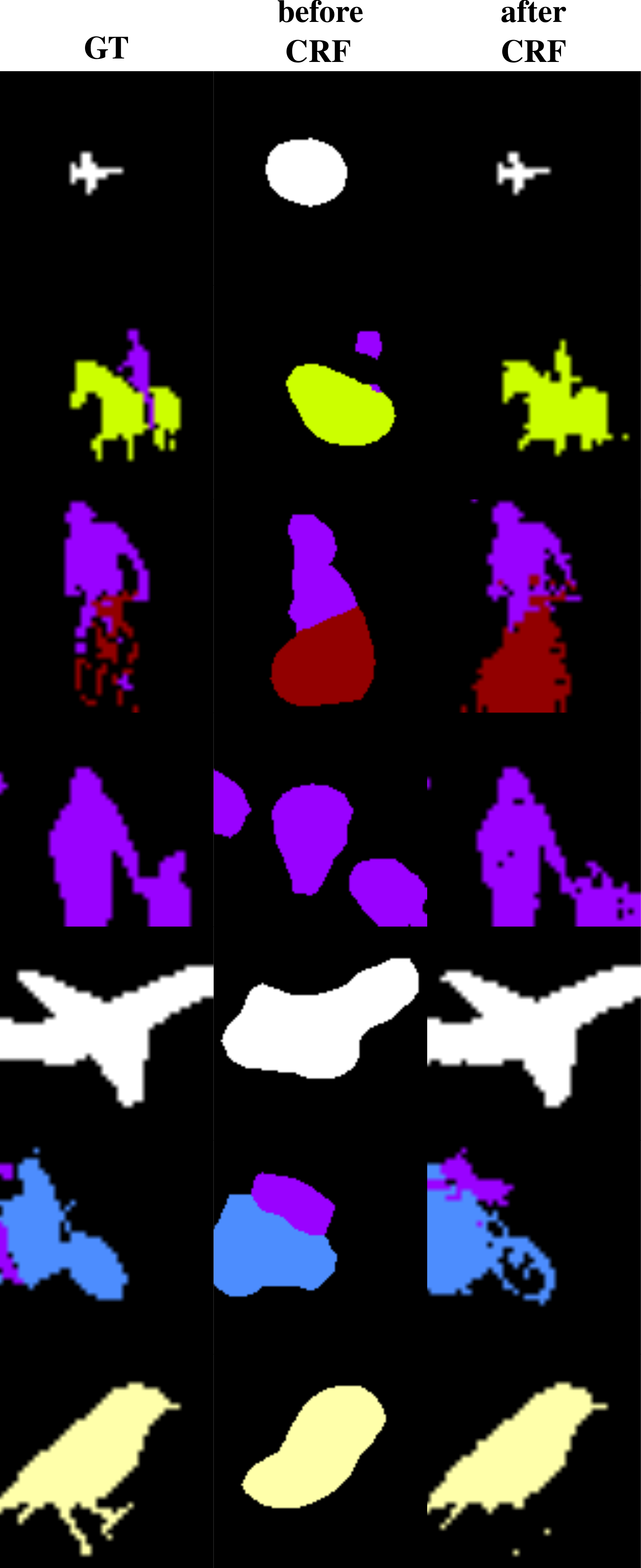}
\caption{Qualitative examples of the effect of using a fully connected CRF model during the learning process. The first column corresponds to the ground truth (GT) masks. The second column shows the learned activation masks before applying our CRF, and the third column shows the masks after the application of the CRF.}
\label{fig:mask_training}
\end{figure}

We now analyze the effect of smoothing out the noisy activation maps obtained by the Hide-and-Seek architecture using the fully connected CRF model \cite{Krahenbuhl2011}. Table \ref{table:result_crf} shows the results achieved by our model with and without the CRF stage during learning. Note that the table reports also the semantic segmentation performance when a CRF is used for the refinement of the final segmentation. The use of a CRF during training increases the accuracy of the model, we have a gain of nearly $4\%$ in terms of mIoU. Figure \ref{fig:mask_training} qualitatively shows the substantial improvement obtained in the activations masks by introducing a fully connected CRF model during the learning process.

\begin{table}[t]
\caption{Impact of smoothing out the noisy activation maps computed during training process. Accuracy comparison on PASCAL VOC 2012 validation set.}
\label{table:result_crf}
\begin{center}
\scalebox{0.65}{
\begin{tabular}{l|cc|cc}
\toprule
\multicolumn{1}{c|}{Models} & \multicolumn{2}{c|}{\textbf{before CRF}} & \multicolumn{2}{c}{\textbf{after CRF}}\\
\multicolumn{1}{c|}{} & PCA ($\%$) & mIoU ($\%$) & PCA ($\%$) & mIoU ($\%$)\\
\midrule
Our architecture & 72.92 & 34.03 & 75.98 & 36.50\\
%\midrule
Our architecture + CRF & \textbf{77.93} & \textbf{38.29} & \textbf{79.76} & \textbf{40.04}\\
\bottomrule
\end{tabular}
}
\end{center}
\end{table}

\subsubsection{Filter for the inference stage}
\label{sec:knowledge}

Table \ref{table:cam_test_filter} reports the results obtained by our model with and without the filtering process detailed in Section \ref{sec:test}. Our filtering strategy is able to incorporate the knowledge of the class activation maps for the refinement of the final segmentations, improving the accuracy of the solution. The improvement is rather humble, which demonstrates that our fully end-to-end learning process itself ends with an architecture that naturally penalizes those classes that are not present in the image. In any case, the increment offered by our filtering mechanism has consistently appeared in our experiments, always improving the final performance of the model.

\begin{table}[t]
\caption{Weakly-supervised semantic segmentation results introducing the proposed test filtering process. Accuracy comparison on PASCAL VOC 2012 validation set.}
\label{table:cam_test_filter}
\begin{center}
\scalebox{0.6}{
\begin{tabular}{l|cc|cc}
\toprule
\multicolumn{1}{c|}{Models} & \multicolumn{2}{c|}{\textbf{before CRF}} & \multicolumn{2}{c}{\textbf{after CRF}}\\
\multicolumn{1}{c|}{} & PCA ($\%$) & mIoU ($\%$) & PCA ($\%$) & mIoU ($\%$)\\
\midrule
Our architecture + CRF & 77.93 & 38.29 & 79.76 & \textbf{40.04}\\
%\midrule
Our architecture + CRF + test filter & \textbf{78.04} & \textbf{38.55} & \textbf{79.79} & \textbf{40.45}\\
\bottomrule
\end{tabular}
}
\end{center}
\end{table}

\subsubsection{Switching Loss Functions}
\label{sec:switch_res}

The switching loss mechanism that we introduce in Section \ref{sec:sl} has to be analyzed experimentally. We here explore the influence of these ideas, reporting the performance of our architecture under three situations: 1) no switching loss approach is used; 2) fixed early switching; and 3) adaptive switching. This experiment can reveal whether we can improve the traditional per pixel softmax cross-entropy loss based learning process used by most of weakly-supervised semantic segmentation models. Table \ref{table:switch} reports all the results.

To analyze the real influence solely due to the losses, we start by analyzing the performance of the proposed architecture considering that no CRF refinement is used. First things first, our results reveal that the application of any of our switching loss strategies always improves the results. With respect to the \textit{fixed early switching} model, we see that the higher the number of iterations we fix to change from the standard cross-entropy loss to the loss using our learned activation maps, the better. This proves our assumption that during the initial iterations, the learned activation maps are noisier and thus hinder the learning process.

Finally, in row 4 of Table \ref{table:switch} we show the benefits of the \textit{adaptive switching} mechanism. By automatically detecting the noisy activation maps iterations, and switching to the cross-entropy loss, we obtain the best performance of 80.55\% and 40.44\% for PCA and mIoU, respectively. If we now apply the CRF refinement step, the PCA and mIoU metrics increase to 80.56\% and 41.89\%.

From now on, we name our best approach, \ie Hide-and-Seek ($CAM_1$ $\&$ $CAM_2$) + CRF for smoothing + test filter + \textit{adaptive switching} + CRF for refinement, simply as $H\&S$.

\begin{table}[t]
\caption{Impact of switching the loss function during learning process. The architecture for all models is $CAM_1$ $\&$ $CAM_2$ + CRF + test filter. Accuracy comparison on PASCAL VOC 2012 validation set.}
\label{table:switch}
\begin{center}
\scalebox{0.8}{
\begin{tabular}{l|cc}
\toprule
\multicolumn{1}{c|}{Learning} & \multicolumn{2}{c}{\textbf{Accuracy}} \\
\multicolumn{1}{c|}{process} & PCA ($\%$) & mIoU ($\%$)\\
\midrule
\emph{Without} Switching Loss Function & 78.04 & 38.55\\
\midrule
Fixed Switching (1000 iters) & 78.93 & 38.55 \\
%\midrule
Fixed Switching (5000 iters) & 79.47 & 39.36 \\
%\midrule
Adaptive Switching  & \textbf{80.55} & \textbf{40.44}\\
\bottomrule
\end{tabular}
}
\end{center}
\end{table}

\subsubsection{Impact of the end-to-end Learning Framework}
\label{sec:impact_endtoend}

Normally, most weakly-supervised systems tend to rely on external components, which are pre-trained, \eg \cite{Bearman2016,Oh2017,Wei2017,Wei2017b}, so, no end-to-end learning of the whole solution is possible.

Our solution can also be trained using a pretrained model for the CAMs. However, one question immediately rises: is it beneficial to train our weakly-supervised segmentation solution in an end-to-end mode? Other solutions using CAM-based localization networks for the weakly-supervision (\eg \cite{Oh2017}), use a pretrained model, which is normally learned in the segmentation dataset itself. Once these models are trained, their parameters are \emph{loaded and frozen}, and only the part of the deep network in charge of the segmentation is learned. In other words, the solutions are trained in two separated steps, without any interaction. We want to demonstrate with this experiment that it is beneficial for both tasks to be learned jointly. Note that we propose to update the model weights of the \emph{whole} architecture following a joint optimization of both classification and segmentation losses, in an end-to-end learning process.

Table \ref{table:end-to-end} summarizes the accuracy achieved by: 1) our $H\&S$ model; 2) our $H\&S$ model but not trained end-to-end (we simply use a pre-trained siamese CAM network) ($H\&S$-\textit{not e2e}); 3) the best pre-trained CAM-based model described in \cite{Oh2017}, named \textit{Seeds}. Again, the obtained results are shown before and after a CRF post-processing step. For all metrics, our end-to-end proposal outperforms our pre-trained CAM-based localization network. Training our model in an end-to-end fashion, we gain $5\%$ on mIoU terms. Interestingly, our siamese CAM architecture trained end-to-end also outperforms the best results presented in \cite{Oh2017} for a GAP-based method, increasing the accuracy by $2.1\%$. 

\begin{table}[t]
\caption{Impact of an end-to-end learning framework. Accuracy comparison on PASCAL VOC 2012 validation set.}
\label{table:end-to-end}
\begin{center}
\scalebox{0.6}{
\begin{tabular}{l|cc|cc}
\toprule
\multicolumn{1}{c|}{Training} & \multicolumn{2}{c|}{\textbf{before CRF}} & \multicolumn{2}{c}{\textbf{after CRF}}\\
\multicolumn{1}{c|}{strategy} & PCA ($\%$) & mIoU ($\%$) & PCA ($\%$) & mIoU ($\%$)\\
\midrule
$H\&S$-\textit{not e2e} & 75.90 & 34.00 & 77.08 & 36.90\\
Seeds \cite{Oh2017} Pre-trained  & -- & 38.70 & -- & 39.80\\
\midrule
$H\&S$  & \textbf{80.55} & \textbf{40.44} & \textbf{80.56} & \textbf{41.89}\\
\bottomrule
\end{tabular}
}
\end{center}
\end{table}

\subsection{Comparison with State-of-the-art Solutions}
\label{sec:state-art}

\begin{figure*}[t]
\centering
\includegraphics[width=0.9\linewidth]{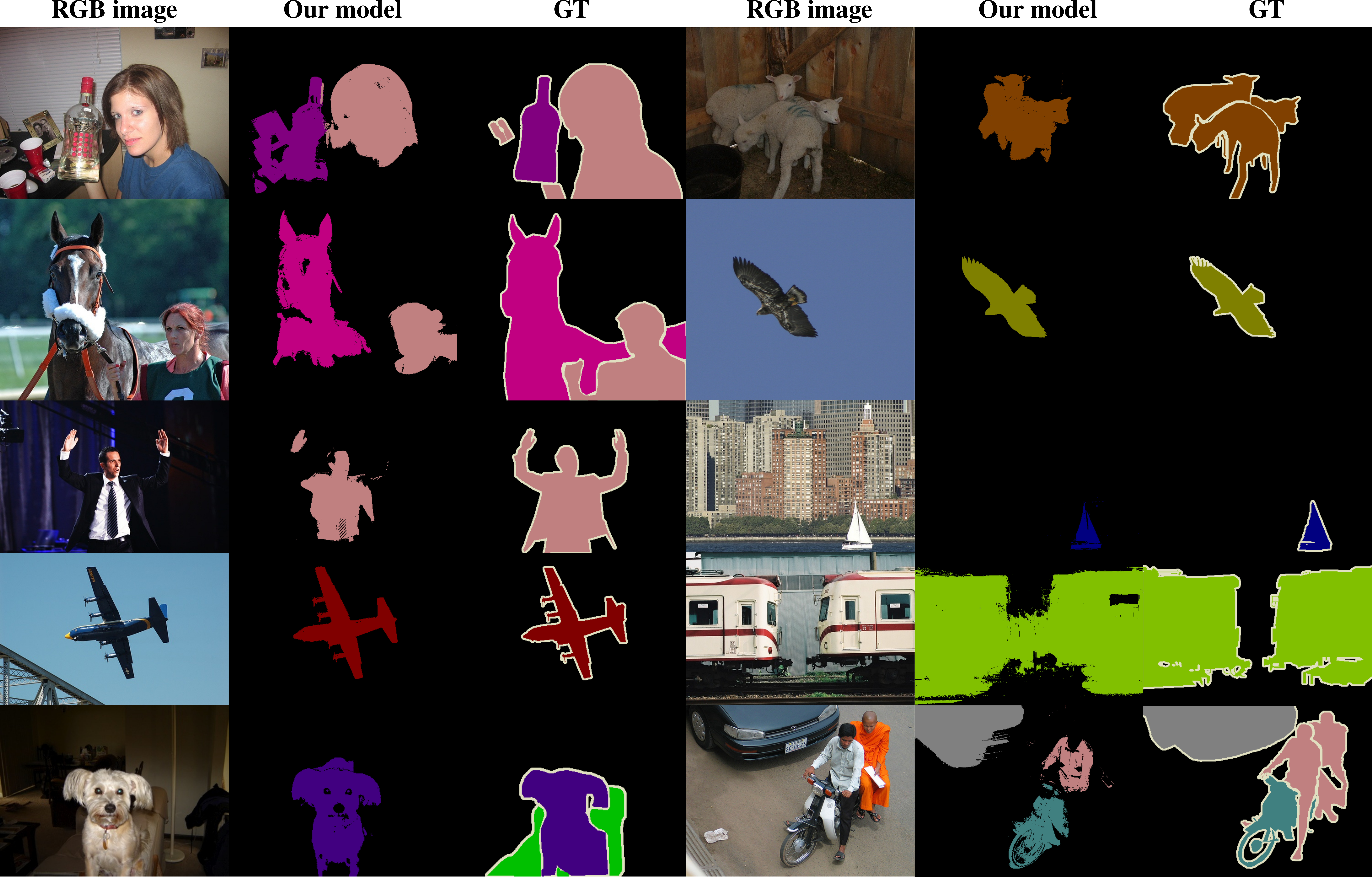}
\caption{Qualitative segmentation results achieved by our $H\&S$ model. The first and forth columns correspond to the RGB images. The second and fifth columns are the masks obtained by our model. Third and sixth columns correspond to ground truth (GT) masks.}
\label{fig:results}
\end{figure*}

We start reporting in Table \ref{table:test_classes} the accuracy, in terms of IoU, achieved by our model for each of the object categories contained in the PASCAL VOC 2012 \cite{voc2012} dataset. Note that our model exceeds $50\%$ of accuracy for classes airplane, bus, car, cat and motorbike, while it presents difficulties to segment the classes bicycle and chair. Qualitative segmentation results are shown in Figure \ref{fig:results}.

\begin{table*}[t]
\caption{Weakly-supervised semantic segmentation results achieved by our $H\&S$ model. Accuracy on 21 classes from PASCAL VOC 2012 validation set.}
\label{table:test_classes}
\begin{center}
\scalebox{0.6}{
\begin{tabular}{l|c|c|c|c|c|c|c|c|c|c|c|c|c|c|c|c|c|c|c|c|c|l}
\toprule
 & Back. & Aero. & Bike & Bird & Boat & Bottle & Bus & Car & Cat & Chair & Cow & Table & Dog & Horse & Mbike & Person & Plant & Sheep & Sofa & Train & Tv & \textbf{Average}\\
\midrule
IoU & 78.14 & 51.32 & 18.20 & 46.55 & 31.92 & 35.08 & 58.52 & 50.60 & 50.24 & 16.45 & 40.33 & 27.43 & 45.13 & 47.85 & 54.87 & 32.05 & 29.51 & 45.97 & 25.74 & 48.68 & 45.12 & \textbf{41.89}\\ 
\bottomrule
\end{tabular}
}
\end{center}
\end{table*}

We now compare our approach with state-of-the-art weakly-supervised solutions trained in an end-to-end fashion only using image-level labels as supervision. Table \ref{table:sota} summarizes the main results for both validation and test sets of the PASCAL VOC 2012 dataset. Note that our approach significantly outperforms 8 of 11 solutions reported in the literature. One exception is the solution of Saleh \etal \cite{Saleh2016}. Note that to reach this result, they need to initialize the weights of the last convolutional layer with the parameters corresponding to the 20 classes shared by PASCAL VOC and ILSVRC datasets. MDC \cite{Wei2018} approach uses a more complex network architecture. They build the approach upon the VGG16 model, removing the fully-connected layers, and one pooling layer, and appending to conv5 layer convoluational blocks with multiple dilated rates. In DSRG \cite{Huang2018}, an external seed region growing mechanism is integrated to refine the activation maps. However, we offer a closed solution, where just the CAMs modules are needed. Finally, for the comparison with GAIN \cite{Li2018}, one must consider that for the results reported, their GAIN mechanism has to be integrated in the more complex SEC model \cite{Kolesnikov2016}. The comparison with the method MIL w/ILP \cite{Pinheiro2015} is particularly interesting. This model uses a large amount of additional images (roughly 700K) from the ILSVRC 2013 dataset, so as to boost the accuracy of the basic MIL method (see row 1). We still outperform this model, even without using any such additional source of data. Finally, the benefits of our class-specific masks are also further evidenced by the fact that we outperform the mask-free models proposed in \cite{Pathak2014} and \cite{Pathak2015} by $16.2$ and $6.6$ mIOU points, respectively.

\begin{table}[t]
\caption{State-of-the-art comparison of weakly-supervised semantic segmentation models trained in an end-to-end manner. Supervision: Image-level Labels. Accuracy comparison on PASCAL VOC 2012 validation and test sets.}
\label{table:sota}
\begin{center}
\scalebox{0.8}{
\begin{tabular}{l|ccc}
\toprule
\multicolumn{1}{c|}{\textbf{Models}} & \multicolumn{1}{c}{\textbf{Training}} & \multicolumn{1}{c}{\textbf{validation}} & \multicolumn{1}{c}{\textbf{test}}\\
\multicolumn{1}{c|}{} & \multicolumn{1}{c}{\textbf{Set}} & \multicolumn{1}{c}{\textbf{(mIoU)}} & \multicolumn{1}{c}{\textbf{(mIoU)}}\\
%\multicolumn{1}{c|}{} & PCA ($\%$) & CCA ($\%$) & IoU ($\%$) & PCA ($\%$) & CCA ($\%$) & IoU ($\%$)\\
\midrule
MIL \cite{Pinheiro2015} & 10K & 17.8 & --\\
EM-Fixed \cite{Papandreou2015} & 10K & 20.8 & --\\
MIL-FCN \cite{Pathak2014}  & 10K & 25.7 & 24.9\\
What's the point \cite{Bearman2016} & 10K & 29.8 & --\\
CCNN \cite{Pathak2015} & 10K & 35.3 & 35.6\\
EM-Adapt \cite{Papandreou2015} & 10K & 38.2 & 39.6\\
BFBP \cite{Saleh2016} & 10K & 46.6 & --\\
GAIN \cite{Li2018} & 10K & 55.3 & 56.8\\
DSRG \cite{Huang2018} & 10K & 59.0 & 60.4\\
MDC \cite{Wei2018} & 10K & 60.4 & 60.8\\
\midrule
$H\&S$ (ours) & 10K & 41.9 & 42.6\\
\midrule
MIL w/ILP \cite{Pinheiro2015} & 700K & 32.6 & --\\
\bottomrule
\end{tabular}
}
\end{center}
\end{table}

To conclude, we consider also interesting to compare the performance of our solution, which only uses image-level labels as supervision, with other methods that rely on additional sources of supervision . In particular, these include the point supervision of \cite{Bearman2016}, models that exploit an objectness prior \cite{Bearman2016, Pinheiro2015}, methods that implicitly use pixel-level supervision \cite{Pinheiro2015, Qi2016, Wei2016}, models that employ labeled bounding boxes \cite{Papandreou2015, Pinheiro2015} or scribbles \cite{Lin2016}, approaches that use saliency maps \cite{Oh2017, Wei2017, Wei2017b} or other kinds of supervision \cite{Pathak2015, Saleh2016}. 

Despite being far from methods such as \cite{Lin2016, Oh2017, Papandreou2015}, note how our accuracy is comparable or even higher than of other methods such as \cite{Bearman2016,Pathak2015, Pinheiro2015}. Importantly, we outperform methods exploiting an objectness prior \cite{Bearman2016, Pinheiro2015}. This fact demonstrates the benefits of using the class-specific segmentation masks provided by our Hide-and-Seek strategy, instead of external and pretrained objectness modules. It is worth mentioning that our model obtains results comparable to \cite{Bearman2016}, where the additional supervision of using annotated points is needed. Using only image tags, our model improves the performance of models that rely on the supervision offered by annotated bounding boxes, as in \cite{Pinheiro2015}, where it is reported a mIoU of $37.0\%$, versus our $42.6\%$ for this metric. We also outperform the random crop supervision implemented in \cite{Pathak2015} ($41.9\%$ vs. $36.4\%$). We achieve an accuracy comparable to that of methods that implicitly use pixel-level supervision \cite{Pinheiro2015, Wei2016} or use size information \cite{Pathak2015}. We believe that this further evidences the benefits of our approach.

\begin{table}[t]
\caption{Accuracy comparison on PASCAL VOC 2012 validation and test sets for other state-of-the-art methods trained with higher levels of supervision.}
\label{table:sota_super}
\begin{center}
\scalebox{0.65}{
\begin{tabular}{l|ccc}
\toprule
\multicolumn{1}{c|}{\textbf{Models}} & \multicolumn{1}{c}{\textbf{Training}} & \multicolumn{1}{c}{\textbf{val}} & \multicolumn{1}{c}{\textbf{test}}\\
\multicolumn{1}{c|}{} & \multicolumn{1}{c}{\textbf{Set}} & \multicolumn{1}{c}{\textbf{(mIoU)}} & \multicolumn{1}{c}{\textbf{(mIoU)}}\\
%\multicolumn{1}{c|}{} & PCA ($\%$) & CCA ($\%$) & IoU ($\%$) & PCA ($\%$) & CCA ($\%$) & IoU ($\%$)\\
\midrule
\textbf{Supervision: Image-level Labels + Points} &  & \\
What's the point + 1Point \cite{Bearman2016} & 10K & 35.1 & --\\
What's the point + obj + 1Point \cite{Bearman2016} & 10K & 42.7 & -- \\
What's the point + obj + AllPoints \cite{Bearman2016} & 10K & 42.7 & -- \\
\midrule
\textbf{Supervision: Image-level Labels + Pixel-level} &  & \\
SN-B+MGG seg \cite{Wei2016}  & 10K & 41.9 & 43.2\\
MIL w/ILP seg \cite{Pinheiro2015} & 700K & 42.0 & 40.6\\
AF-MGG seg \cite{Qi2016} & 10K & 54.3 & 55.5\\
\midrule
\textbf{Supervision: Image-level Labels + Boxes} &  & \\
MIL w/ILP + bbox \cite{Pinheiro2015} & 700K & 37.8 & 37.0\\
WSSL + bbox \cite{Papandreou2015} & 10K & 60.6 & 62.2\\
\midrule
\textbf{Supervision: Image-level Labels + Scribbles} &  & \\
Scribblesup \cite{Lin2016} & 10K & 63.1 & --\\
\midrule
\textbf{Supervision: Image-level Labels + Others} &  & \\
CCNN + Random Crops \cite{Pathak2015} & 10K & 36.4 & --\\
CCNN + Size Info \cite{Pathak2015}  & 10K & 42.4 & 45.1\\
BFBP + CheckMask \cite{Saleh2016} & 10K & 51.5 & 52.9\\
\midrule
\textbf{Supervision: Image-level Labels + Saliency} &  & \\
STC \cite{Wei2017b} & 50K & 49.8 & 51.2\\
AE-PSL \cite{Wei2017} & 10K & 55.0 & 55.7\\
$\mathcal{G}_2$ \cite{Oh2017} & 10K & 55.7 & 56.7\\
\midrule
\textbf{Supervision: Image-level Labels + Objectness} &  & \\
What's the point \cite{Bearman2016} & 10K & 32.2 & --\\
MIL w/ILP-sppxl \cite{Pinheiro2015} & 700K & 36.6 & 35.8\\
\midrule
\textbf{Supervision: Image-level Labels} &  & \\
GAIN \cite{Li2018} & 10K & 55.3 & 56.8\\
DSRG \cite{Huang2018} & 10K & 59.0 & 60.4\\
MDC \cite{Wei2018} & 10K & 60.4 & 60.8\\
$H\&S$ (ours) & 10K & 41.9 & 42.6\\
\bottomrule
\end{tabular}
}
\end{center}
\end{table}

\subsection{Weakly Supervised Object Localization}
\label{sec:wsol}

Do object detectors emerge in our architecture? Although weakly supervised object localization (WSOL) is not the main target of our research, it is true that our Hide-and-Seek block, just learning from image labels, is able to produce object localizations. Therefore, we evaluate here the performance for this task, using again the PASCAL VOC 2012 dataset, which also provides ground truth information for the bounding boxes of the objects.

Technically, we compute the class activation maps (CAMs) released by the Hide-and-Seek module for each image in the validation set of the PASCAL VOC 2012 dataset. Our objective is to create a set of bounding boxes (BBs) for each category. Therefore, we start performing a binarization to the CAMs, using a threshold of 0.6. This means that regions in the CAMs with values over this threshold do contain objects. Then, we apply a standard connected components algorithm to consolidate the regions, and build the final BBs that enclose them. The score of each BB is computed as the sum of the output scores in the CAM associated to the corresponding regions. As a final refinement, we perform a Non-max suppression.

\begin{figure*}
\centering
\includegraphics[width=0.9\linewidth]{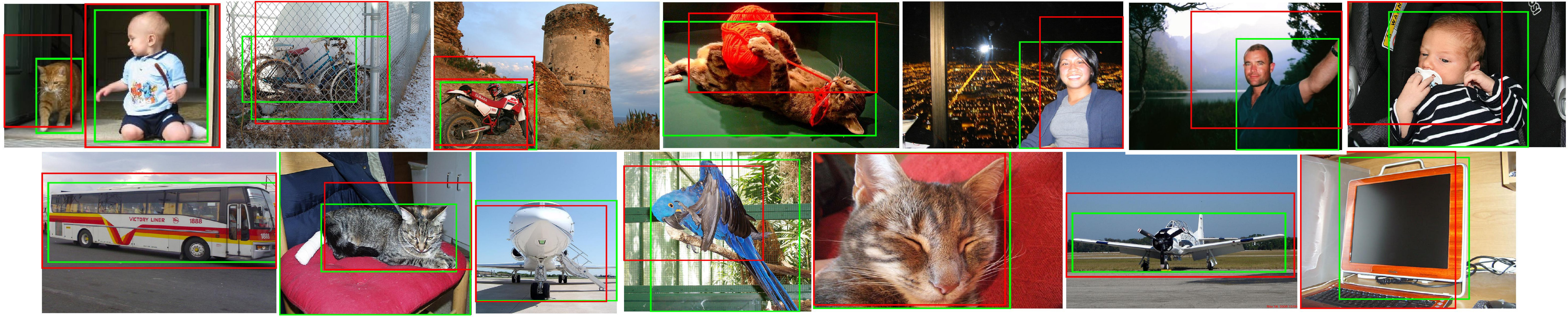}
\caption{Qualitative Weakly Supervised Object Localization results achieved by our $H\&S$ model. Green boxes indicate ground-truth instance annotation. Red boxes indicate our localizations.}
\label{fig:qualitative_wsol}
\end{figure*}

We report in Table \ref{table:wsol} the mean Average Precision (mAP) for different intersection over union thresholds (tIoU), as suggested in \cite{Hoiem2012}. Interestingly, for categories aeroplane, bus, cat, dog, motorbike and train, our approach reports a high Average Precision (AP). Figure \ref{fig:qualitative_wsol} shows some qualitative results for the WSOL problem.

\begin{table}[b]
\caption{Weakly Supervised Object Localization results achieved by our $H\&S$ model. Accuracy on 20 classes from PASCAL VOC 2012 validation set. The Average Precision per category is reported.}
\label{table:wsol}
\begin{center}
\scalebox{0.8}{
\begin{tabular}{l|c c}
\toprule
 AP & tIoU (0.5) & tIoU (0.2) \\
 Aeroplane & 29.3& 62.7\\
 Bike & 26.1 & 50.5\\
 Bird & 3.8 & 29.3\\
 Boat & 4.3 & 17.0\\
 Bottle & 1.5 & 9.2\\
 Bus & 30.9 & 56.6\\
 Car & 6.9 & 23.5\\
 Cat & 26.3 & 73.7\\
 Chair & 0.7 & 10.0\\
 Cow & 4.1 & 18.8\\
 Table & 10.7 & 39.0\\
 Dog & 13.8 & 59.4\\
 Horse & 11.7 & 42.4\\
 Mbike & 27.2 & 52.0\\
 Person & 4.9 & 29.6\\
 Plant & 3.6 & 9.1\\
 Sheep & 3.5 & 12.8\\
 Sofa & 7.0 & 34.5\\
 Train & 18.1 & 59.4\\
 Tv & 4.5 & 29.8\\
 \midrule
\textbf{Average} & \textbf{11.94} & \textbf{35.96}\\
\bottomrule
\end{tabular}
}
\end{center}
\end{table}

% \begin{table}[b]
% \caption{Weakly Supervised Object Localization results achieved by our $H\&S$ model. Accuracy on 20 classes from PASCAL VOC 2012 validation set. The Average Precision per category is reported.}
% \label{table:wsol}
% \begin{center}
% \scalebox{0.6}{
% \begin{tabular}{l|c c c c c c c c c c c c c c c c c c c c|l}
% \toprule
%  &  Aero. & Bike & Bird & Boat & Bottle & Bus & Car & Cat & Chair & Cow & Table & Dog & Horse & Mbike & Person & Plant & Sheep & Sofa & Train & Tv & \textbf{Average}\\
% 
% \midrule
% tIoU (0.5)  & 29.3 & 26.1 & 3.8 & 4.3 & 1.5 & 30.9 & 6.9 & 26.3 & 0.7 & 4.1 & 10.7 & 13.8 & 11.7 & 27.2 & 4.9 & 3.6 & 3.5 & 7.0 & 18.1 & 4.5 & 
% \textbf{11.94}\\ 
% \midrule
% tIoU (0.2)  & 62.7 & 50.5 & 29.3 & 17.0 & 9.2 & 56.6 & 23.5 & 73.7 & 10.0 & 18.8 & 39.0 & 59.4 & 42.4 & 52.0 & 29.6 & 9.1 & 12.8 & 34.5 & 59.4 & 29.8 & 
% \textbf{35.96}\\ 
% \bottomrule
% \end{tabular}
% }
% \end{center}
% \end{table}

\section{Conclusion}
\label{sec:conclusion}
We have addressed the problem of weakly-supervised semantic segmentation using only image-level labels. In particular, we have proposed an end-to-end learning method to directly exploit the prior knowledge of a CNN network, trained for object recognition, to generate class-specific masks, which are used as automatic supervisory mechanisms to train the segmentation network.

In particular, we have introduced a novel siamese CNN architecture, \ie the Hide-and-Seek, based on the CAM technique, which learns to generate class specific activation maps that are able to cover the full object extents. We have also shown how to integrate these multi-class masks into a new end-to-end learning process, which allows a CNN architecture to jointly learn to classify and segment the images. Our experiments on PASCAL VOC 2012 \cite{voc2012} dataset have shown the benefits of our approach, which outperforms several methods that use image tags only, and even some models that leverage additional supervision or training data.

\section*{Acknowledgments}
This work is supported by project PREPEATE, with reference TEC2016-80326-R, of the Spanish Ministry of Economy, Industry and Competitiveness. We gratefully acknowledge the support of NVIDIA Corporation with the donation of a GPU used for this research. Cloud computing resources were kindly provided through a Microsoft Azure for Research Award.

\ifCLASSOPTIONcaptionsoff
  \newpage
\fi



\bibliographystyle{IEEEtran}
\bibliography{egbib}
% \end{thebibliography}


\begin{IEEEbiography}[{\includegraphics[width=1in,height=1.25in,clip,keepaspectratio]{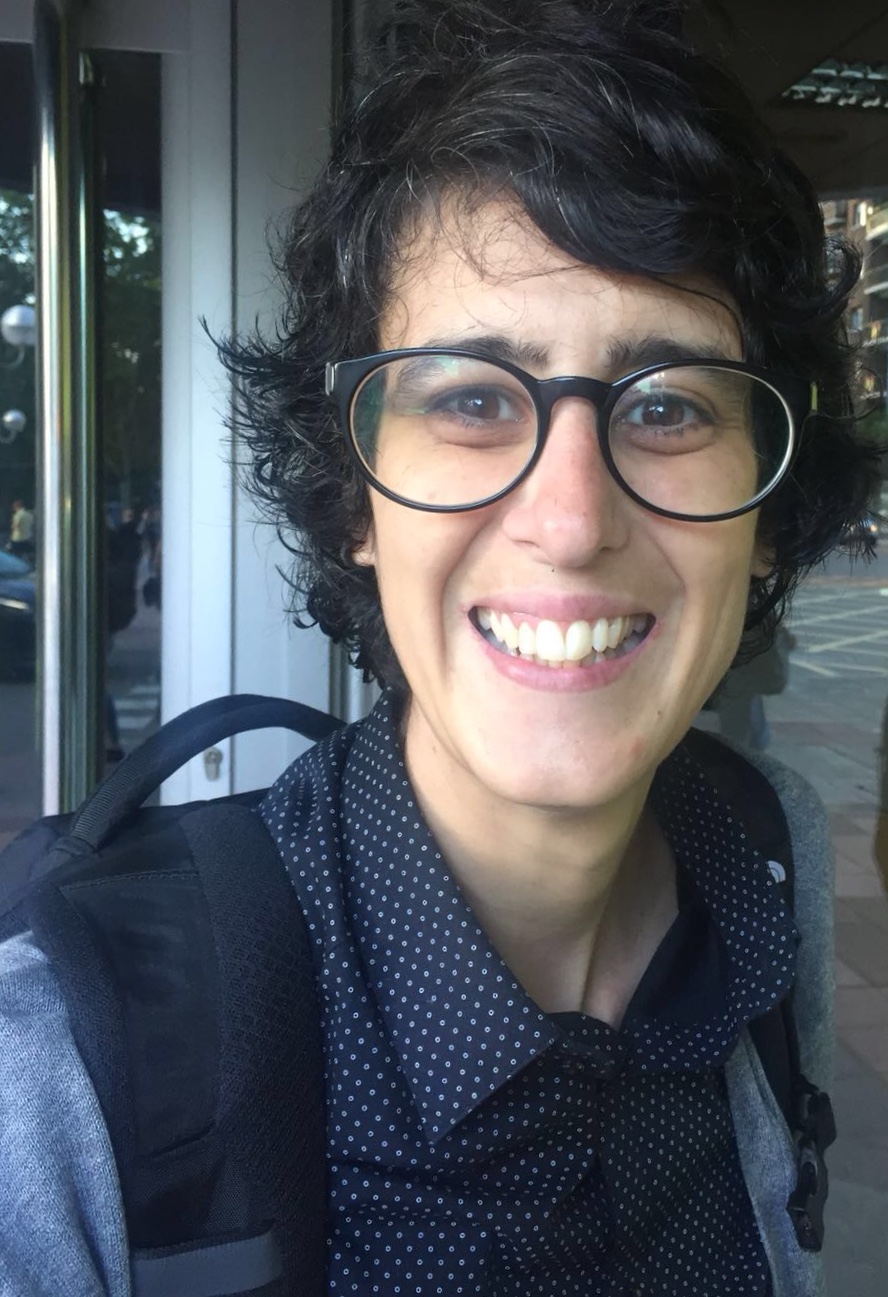}}]{Carolina Redondo-Cabrera} was born on January 1st, 1983 in Toledo city, Spain. She received her Bachelor of Science (B.Sc), in field of Telecommunications, from the University of Alcal\'a, Spain, in 2009. She completed her Master of Science (Msc), in the field of ICT, from the University of Alcal\'a in 2012 and then joined the GRAM research group (University of Alcal\'a) to continue her study to Doctor of Philosophy (PhD) in the field of Computer Vision. Her research interests are centered around computer vision, machine learning and deep learning, focusing on object detection and image understanding.
\end{IEEEbiography}
\begin{IEEEbiography}[{\includegraphics[width=1in,height=1.25in,clip,keepaspectratio]{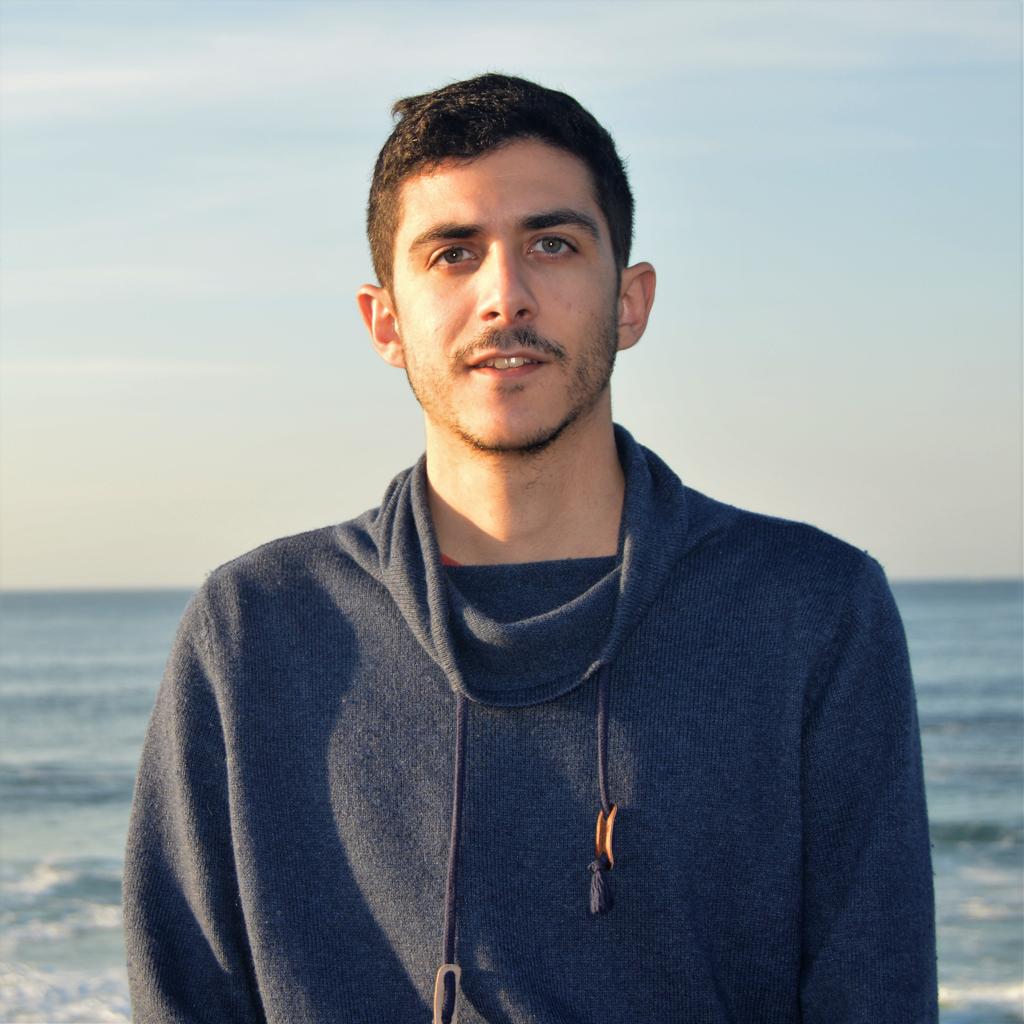}}]{Marcos Baptista-R\'ios} is currently a Ph.D. student at Universidad de Alcal\'a (Spain) in GRAM Vision Lab research group. His research interest is manly focused on Deep Learning for Computer Vision. More specifically, his work is focused on video understanding and activity recognition.
\end{IEEEbiography}
\begin{IEEEbiography}[{\includegraphics[width=1in,height=1.25in,clip,keepaspectratio]{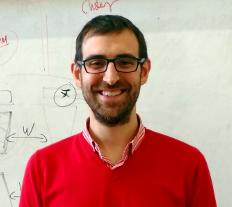}}]{Roberto L\'opez-Sastre}
is an Associate Professor in the Department of Signal Theory and Communications of the University of Alcal\'á. His research interests are centered around computer vision, machine learning and robotics, focusing on object detection, scene understanding and action recognition. He serves as program committee member/reviewer of the major computer vision conferences ICCV, ECCV, and CVPR.
\end{IEEEbiography}


% % insert where needed to balance the two columns on the last page with
% %\newpage


%\vfill


\end{document}